\documentclass[lettersize,journal]{IEEEtran}
\usepackage{amsmath,amsfonts}
\usepackage{algorithmic}
\usepackage{algorithm}
\usepackage{array}
\usepackage{subfig}
\usepackage[colorlinks,
linkcolor = blue,
anchorcolor = blue,
citecolor = blue,
]{hyperref}
\usepackage{textcomp}
\usepackage{stfloats}
\usepackage{url}
\usepackage{verbatim}
\usepackage{graphicx}
\usepackage{cite}
\usepackage{enumitem}
\usepackage{multirow}
\usepackage{multicol}
\usepackage{booktabs}
\usepackage{makecell}
\usepackage{bm}
\usepackage{color} 
\usepackage{soul}
\usepackage{mathtools}
\usepackage{textcomp}
\usepackage{gensymb}
\usepackage{enumitem}
\usepackage{amssymb}
\usepackage{pifont}
\usepackage{amsthm}
\usepackage{multirow}
\usepackage{rotating}
\usepackage{adjustbox}
\usepackage{threeparttable}
\usepackage[dvipsnames]{xcolor}
\usepackage{bbding}
\usepackage[utf8]{inputenc}   % �� pdfLaTeX

\usepackage{url}

\begin{document}

%\title{A New Method for Optimizing the Comprehensive Performance of Crowd Navigation from a Trajectory Perspective}
\title{A New Trajectory-Oriented Approach to Enhancing Comprehensive Crowd Navigation Performance}
%\section{A Novel Trajectory-Based Approach to Optimizing the Comprehensive Performance of Social Navigation}

\author{Xinyu Zhou, Songhao Piao, Chao Gao, and Liguo Chen 

\thanks{Xinyu Zhou and Songhao Piao are with the xxx (e-mail: zhouxy@stu.hit.edu.cn; xxx).}
\thanks{Chao Gao are with xxx, China (e-mail: xxx).}
\thanks{Liguo Chen are with xxx, China (e-mail: xxx).}

\thanks{Corresponding author: Songhao Piao and Chao Gao}
}

\maketitle

\begin{abstract} 

%Crowd navigation has garnered significant research attention in recent years, particularly with the advent of deep reinforcement learning(DRL)-based methods. Current DRL-based approaches primarily aim to improve the efficiency and proximal comfort of social navigation. However, many existing studies have either overlooked trajectory optimization altogether or adopted relatively simple, empirically unvalidated smoothness reward terms, even though effective optimization can greatly enhance naturalness, comfort, and energy efficiency in navigation. To address the existing research gap, this paper explores the correlation between efficiency, comfort, and trajectory quality, and introduces a novel evaluation framework that enables precise assessment of trajectory quality and systematic evaluation of overall performance. Additionally, we propose a new reward shaping method that focuses on optimizing trajectory curvature, aiming to further enhance trajectory quality and improve adaptability across multi-scale scenarios, while ensuring that obstacle avoidance performance does not degrade. Through experimental comparisons, both overall performance and trajectory quality surpass those of current state-of-the-art methods, particularly in terms of C$^2$ continuity. The source code will be publicly released on https://github.com/Zhouxy-Debugging-Den/crowdtraj.

Crowd navigation has garnered considerable research interest in recent years, especially with the proliferating application of deep reinforcement learning (DRL) techniques. Many studies, however, do not sufficiently analyze the relative priorities among evaluation metrics, which compromises the fair assessment of methods with divergent objectives.  Furthermore, trajectory-continuity metrics, specifically those requiring $C^2$ smoothness, are rarely incorporated. Current DRL approaches generally prioritize efficiency and proximal comfort, often neglecting trajectory optimization or addressing it only through simplistic, unvalidated smoothness reward. Nevertheless, effective trajectory optimization is essential to ensure naturalness, enhance comfort, and maximize the energy efficiency of any navigation system. To address these gaps, this paper proposes a unified framework that enables the fair and transparent assessment of navigation methods by examining the prioritization and joint evaluation of multiple optimization objectives. We further propose a novel reward-shaping strategy that explicitly emphasizes trajectory-curvature optimization. The resulting trajectory quality and adaptability are significantly enhanced across multi-scale scenarios. Through extensive 2D and 3D experiments, we demonstrate that the proposed method achieves superior performance compared to state-of-the-art approaches. The source code will be released at \url{https://github.com/Zhouxy-Debugging-Den/crowdtraj}.
%To address these gaps, this paper investigates the prioritization and comprehensive evaluation of multiple optimization objectives and introduces a novel evaluation framework that enables fair and transparent assessment of both individual and overall performance. 
%In addition, we propose a new reward-shaping method that emphasizes trajectory curvature optimization to improve trajectory quality and adaptability across multi-scale scenarios. 
%The indirect motion direction constraint further enhances motion stability, leading to significant overall performance improvements. 
%Extensive 2D and 3D experiments demonstrate that the proposed method outperforms state-of-the-art approaches. 
%The source code will be released at \url{https://github.com/Zhouxy-Debugging-Den/crowdtraj}.
%

%Additionally, an indirect motion-direction constraint is integrated to boost motion stability, yielding notable gains in overall performance.
\end{abstract}


\section{Introduction}
\label{sec:1}

%\IEEEPARstart{I}{n} recent years, socially aware robot navigation has gradually attracted people's attention and has become an independent field of research\cite{kruse2013human}, \cite{charalampous2017recent}, \cite{singamaneni2024survey}. With the increasing number of service robots and autonomous vehicles, their ability to efficiently and seamlessly perform navigation tasks around humans is crucial. Appropriate social behavior is key for robots to gain human acceptance and avoid causing any discomfort.

%\IEEEPARstart{I}{n} recent years, crowd navigation has gained increasing attention and evolved into an independent research field~\cite{kruse2013human,charalampous2017recent,singamaneni2024survey}. 
%With the growing deployment of service robots and autonomous vehicles, the ability to navigate efficiently and seamlessly in human environments has become essential. 
%Exhibiting appropriate social behavior is critical for ensuring human acceptance and preventing discomfort.

\IEEEPARstart{C}{rowd} navigation has attracted increasing attention in recent years, developing into a distinct research domain \cite{kruse2013human},\cite{charalampous2017recent},\cite{singamaneni2024survey}. With the wider deployment of service robots and autonomous vehicles, efficient, seamless, and socially appropriate operation in human environments is essential for ensuring human acceptance and avoiding discomfort.

Early work in crowd navigation primarily categorized methods into model-based and trajectory-based approaches. Model-based methods (e.g., ORCA \cite{van2011reciprocal} and SFM \cite{helbing1995social}) define navigation using geometric or mechanical formulations. While conceptually straightforward, their short-sighted nature makes them prone to the “reciprocal dance” phenomenon \cite{feurtey2000simulating} and requires extensive hand-crafted tuning for generalization. Trajectory-based approaches, in contrast, typically adopt a predict-then-plan paradigm \cite{chen2020relational,cao2019dynamic}. Though they mitigate some model-based limitations, they often remain overly conservative, frequently leading to the “freezing robot problem” \cite{trautman2010unfreezing}.

%Deep Reinforcement Learning (DRL) provides an alternative approach to implicitly integrate interactive prediction and planning, shifting the heavy online computation to offline training. Numerous studies have explored this direction based on DRL \cite{shi2022enhanced}, \cite{zhu2022collision}, \cite{yang2023st}.

%Deep Reinforcement Learning (DRL) offers an alternative approach that implicitly integrates interaction prediction and planning, shifting intensive online computation to offline training. 
%Numerous studies have explored this direction~\cite{shi2022enhanced,zhu2022collision,yang2023st}.
%However, evaluating crowd navigation performance remains a challenging task. 
%First, existing studies lack quantitative indicators for trajectory smoothness. 
%Second, there is no unified evaluation framework capable of comprehensively assessing performance across multiple objectives, including safety, success rate, comfort, trajectory quality, and efficiency. This lack of a unified standard makes it difficult to objectively evaluate methods with differing strengths. 
%For example, some approaches excel in success rate, while others perform better in trajectory quality, making it challenging to determine which method is superior overall.

Deep reinforcement learning (DRL) offers an alternative paradigm that implicitly integrates interaction prediction and planning, shifting most computational effort from online execution to offline training. Although this direction has been widely explored~\cite{shi2022enhanced},~\cite{zhu2022collision},~\cite{yang2023st}, evaluating crowd-navigation performance remains challenging. Quantitative metrics for trajectory smoothness are largely absent, and no unified framework exists for jointly assessing safety, success rate, comfort, trajectory quality, and efficiency. Consequently, methods with different strengths—for example, one excelling in success rate and another in trajectory quality—are difficult to compare fairly and holistically.

%Meanwhile, existing DRL-based methods provide limited optimization for trajectory generation and are often unsuitable for holonomic robots. 
%In addition, little attention has been paid to improving $C^2$ trajectory continuity, which not only affects trajectory smoothness but also influences subsequent decision-making. 
%Enhancing trajectory continuity reduces local oscillations, constrains abrupt velocity changes, and stabilizes motion during obstacle avoidance. 
%This simplification of motion control improves obstacle avoidance performance, increases overall success rate, and also enhances pedestrian comfort, as illustrated in Fig. \ref{fig:traj_to_comfort}. Moreover, smoother trajectories generally result in more energy-efficient motion.

%Meanwhile, existing DRL-based methods offer limited support for trajectory optimisation and are often ill-suited to holonomic robots. Moreover, little attention has been given to improving $C^2$ trajectory continuity, which influences not only smoothness but also subsequent decision-making. Enhancing continuity helps suppress local oscillations, restrict abrupt velocity changes, and stabilise motion during obstacle avoidance. This simplification of motion control improves avoidance performance, raises overall success rates, and enhances pedestrian comfort, as illustrated in Fig.~\ref{fig:traj_to_comfort}. Furthermore, smoother trajectories typically yield greater energy efficiency.

\begin{figure}[t]
	\begin{center}
		\includegraphics[width=0.75\linewidth]{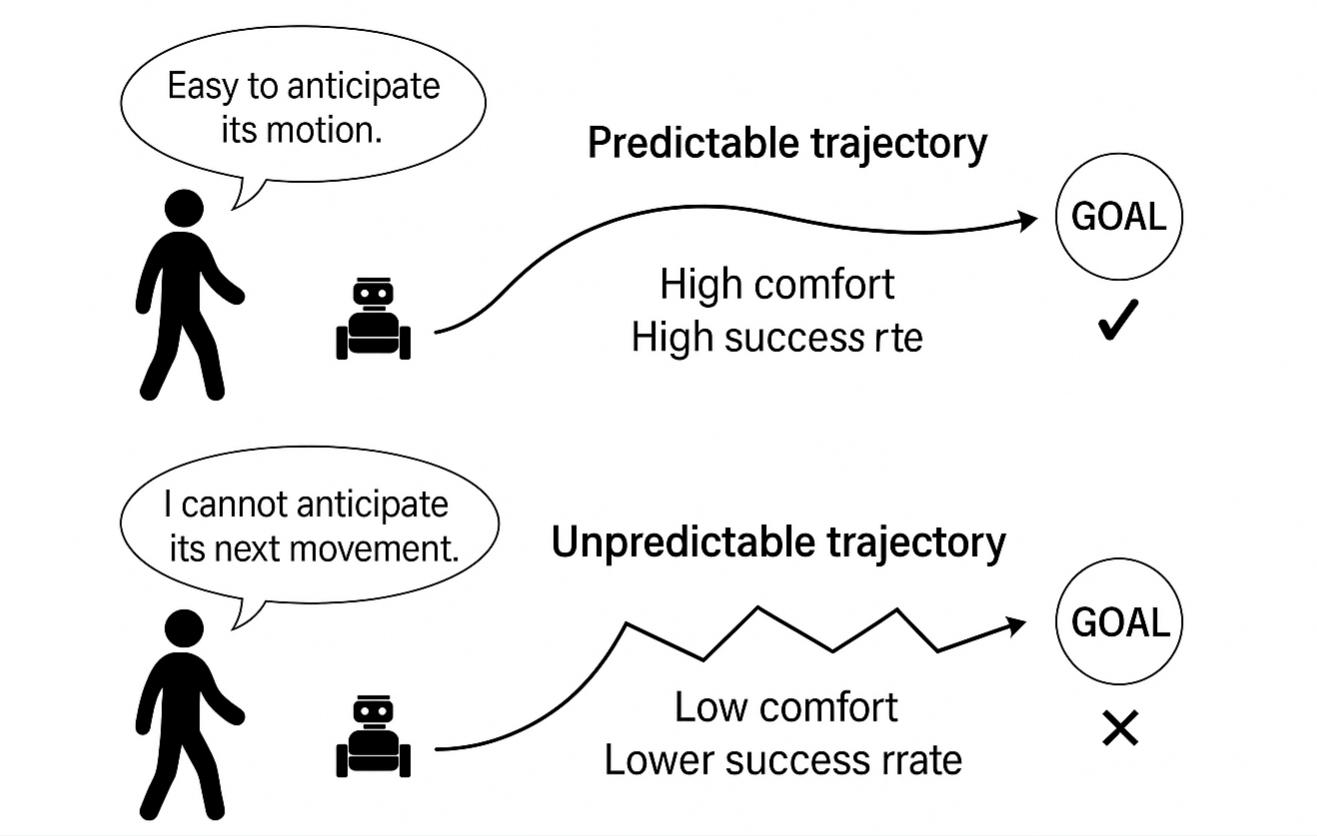}
	\end{center}
	\captionsetup{font=small}
	\caption{Impact of trajectory continuity on crowd navigation.}
	\label{fig:traj_to_comfort}
\end{figure}

Meanwhile, existing DRL-based methods offer limited support for trajectory optimization and are often ill-suited to holonomic robots. Moreover, little attention has been given to improving $C^2$ trajectory continuity, a critical factor that influences both smoothness and subsequent decision-making. Enhancing continuity suppresses local oscillations, restricts abrupt velocity changes, and stabilizes motion during obstacle avoidance. This simplification of motion control improves avoidance performance, raises overall success rates, and enhances pedestrian comfort, as illustrated in Fig.~\ref{fig:traj_to_comfort}. Furthermore, smoother trajectories inherently yield greater energy efficiency.

Accordingly, this study proposes a multi-objective evaluation framework to address these limitations. We further introduce a noval reward-shaping strategy that quantitatively enhances trajectory continuity and improves overall navigation performance.

%Furthermore, the common initialization with imitation learning from ORCA \cite{van2011reciprocal} often propagates ORCA’s inherent limitations, thereby constraining subsequent training. 
%To address these issues, we replace imitation learning with dense rewards, which promotes more stable convergence. This design enables smoother and more predictable robot motion, enhancing pedestrian comfort, while simultaneously reducing low-frequency speed bursts to lower energy consumption (Fig.~1). Consequently, our work emphasizes improving trajectory smoothness without sacrificing efficiency and proximity comfort, offering a more balanced solution for safe and natural robot navigation, 

%There are not only many challenges in implementation, but there are also many challenges in evaluation. It's not hard to see that social navigation is a continuous decision-making problem with multiple objectives. ``Social conformity'' is a type of soft constraint that comprises many aspects, including safety, comfort, naturalness, and other social preferences\cite{singamaneni2024survey}, \cite{mavrogiannis2023core}. This poses a number of challenges, such as how to evaluate the overall performance of pedestrian navigation strategies in terms of efficiency, proximity comfort, and trajectory quality, for example, and whether methods can be explored to improve trajectory quality without sacrificing efficiency and proximity comfort. Therefore, the main research in this paper focuses on how to design a more reasonable comprehensive evaluation index for pedestrian navigation, and at the same time innovatively design a reward shaping method, so that the strategy not only improves the efficiency and proximity comfort, but also greatly improves the quality of trajectory under the continuous action space. The primary contributions are outlined as follows.

\begin{itemize}[topsep=0.0cm, itemsep=0.0cm, parsep=0.0cm]
	\item This study introduces a priority-based evaluation framework for multi-objective optimization, explicitly clarifying the relationships between individual objectives and overall performance.
	\item In crowd navigation, this study introduces a new metric for trajectory continuity, specifically designed to assess $C^2$-level smoothness.
	\item We propose a reward-shaping method that markedly improves the trajectory quality of holonomic robots using DRL-based approaches in high-density environments. With minimal compromise in comfort, it substantially enhances safety, success rate, and other key metrics, thereby yielding a significant overall performance gain.
\end{itemize}

The remainder of this paper is structured as follows: Section \ref{sec:2} reviews related work; Section \ref{sec:3} presents the theoretical formulation; Section \ref{sec:4} details the experimental design and analyses the results; and Section \ref{sec:5} concludes the paper.

\section{Related Work}
\label{sec:2}

\subsection{DRL-based crowd navigation using discrete actions}
\label{sec:2.1}

Since the introduction of CADRL by Chen \textit{et al.}~\cite{chen2017decentralized}, reinforcement learning has become a central tool for collision avoidance, often outperforming traditional methods such as ORCA. However, its fixed input dimension limits adaptability in multi-agent scenarios. Everett \textit{et al.}~\cite{everett2018motion} addressed this limitation in GA3C-CADRL using an LSTM, though human–human interaction (HHI) was not modeled. Chen \textit{et al.}~\cite{chen2019crowd} introduced SARL, employing self-attention to capture both human–robot interaction (HRI) and HHI, but with a still coarse representation of the latter. Liu \textit{et al.}~\cite{liu2021decentralized} incorporated spatio-temporal interactions through DSRNN, yet temporal modeling of HHI remained limited. More recent transformer-based approaches, such as ST$^2$~\cite{yang2023st}, demonstrate strong capability in encoding spatio-temporal dependencies in HRI and HHI. Nevertheless, they generally overlook trajectory quality and action continuity, often relying on coarse discrete action spaces that hinder fine-grained optimization. Although Liu \textit{et al.}~\cite{liu2023intention} enriched the social graph with intent information and temporal edges and adopted a continuous action space, trajectory continuity itself was not explicitly optimized.

\subsection{Trajectory Optimization in Traditional Navigation}
\label{sec:2.2}

\begin{figure}[t]
	\begin{center}
		\includegraphics[width=0.85\linewidth]{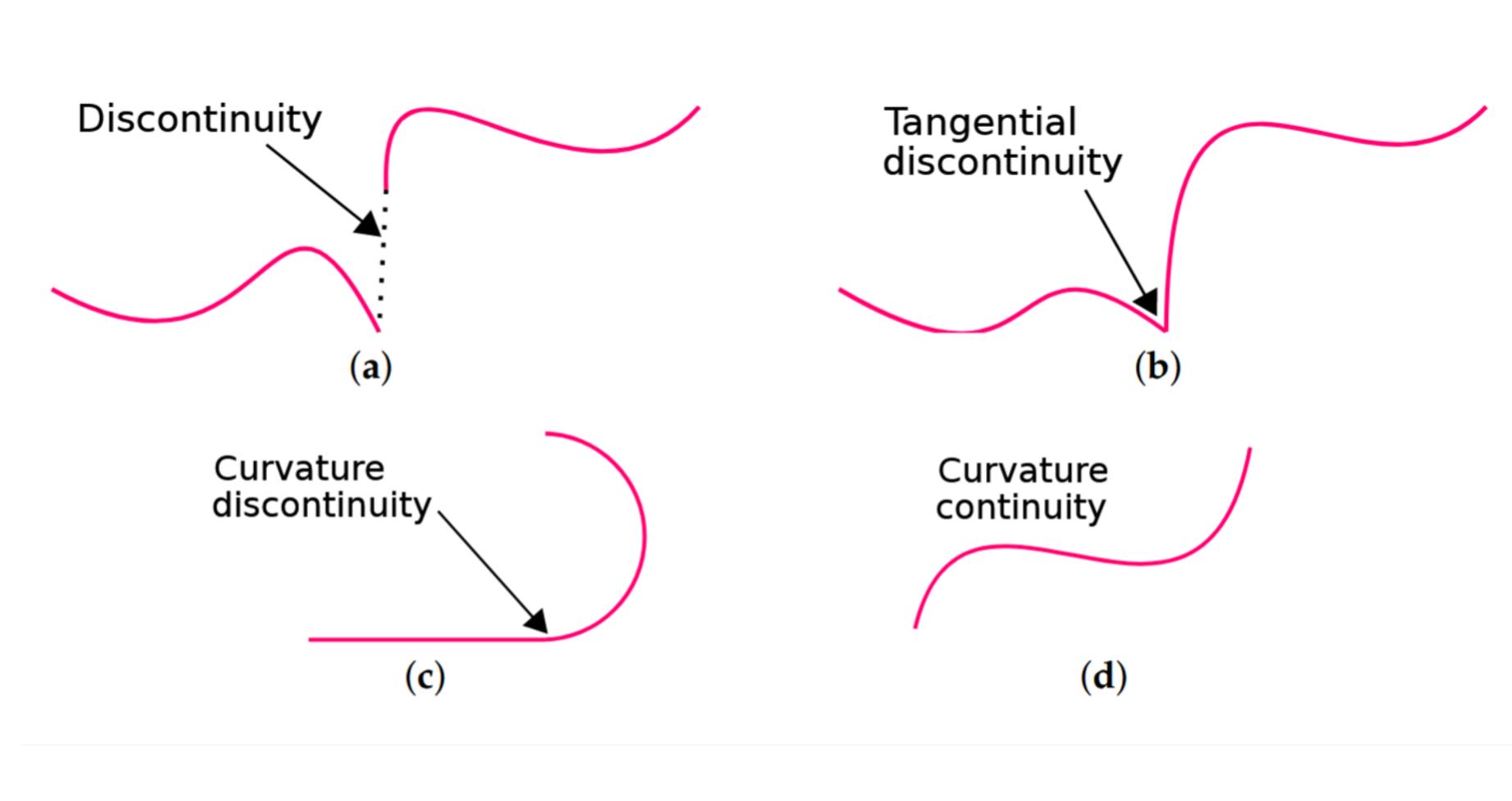}
	\end{center}
	\captionsetup{font=small}
	\caption{Schematic diagram of trajectories with different levels of continuity~\cite{ravankar2018path}. (a) Discontinuous trajectory. (b) $C^0$ continuity. (c)$C^1$ continuity. (d) $C^2$ continuity. }
	\label{fig: continuity}
\end{figure}

Traditional trajectory-generation methods improve smoothness but often lack curvature-level guarantees. Methods like Dubins curves~\cite{yang20132d} and Shortest Homotopic Paths (SHP)~\cite{ravankar2016shp} provide only $C^{1}$ continuity, which may suffice at low speeds but becomes inadequate for high-speed or safety-critical manoeuvres. Continuity distinctions are illustrated in Fig.~\ref{fig: continuity}. Hybrid approaches that combine geometric or spline-based planners with dynamic optimisation~\cite{rosmann2012trajectory} enhance obstacle avoidance but rely on hand-crafted objectives, limiting generalisation and incurring high computational cost in dynamic environments. To address these issues, this study proposes a DRL-based trajectory-generation framework that explicitly enforces $C^{2}$ continuity (see Fig.~\ref{fig: continuity}), combining the adaptability of learning-based planning with curvature-smooth, dynamically feasible trajectories suited to dense and rapidly evolving scenarios.

\label{fig:3}

\subsection{Reward Shaping for Smoothing in DRL-based crowd navigation}
\label{sec:2.3}

Several studies have introduced reward and penalty terms to encourage trajectory smoothness, often noting its close relation to angular-velocity correlation. DenseCAvoid~\cite{sathyamoorthy2020densecavoid}, for example, penalises abrupt angular-velocity changes and incorporates waypoint guidance to reduce oscillatory behaviour. MODSRL~\cite{cheng2024multi} employs a multi-reward formulation that balances safety, efficiency, collision avoidance, and path smoothness, with angular-velocity threshold penalties acting as key smoothing components. Other works penalise acceleration: Where to Go Next~\cite{brito2021go} uses an MPC-based reward that penalises both linear and angular acceleration, while NAX~\cite{matsuzaki2022learning} jointly minimises angular velocity and jerk to improve smoothness.
Despite these varied reward-shaping strategies, existing approaches lack quantitative evaluation of trajectory quality and do not explicitly identify which aspects of trajectory continuity are improved.

\section{Methodology}
\label{sec:3}

\subsection{Problem Statement}

\subsubsection{Problem Formulation of Crowd Navigation in Deep Reinforcement Learning}
\label{sec:3.2}

%Due to the limited sensing range of onboard sensors, the crowd navigation problem is typically modeled as a partially observable Markov decision process (POMDP), defined by the tuple $\langle \mathcal{S}, \mathcal{A}, \mathcal{P}, \mathcal{R}, \gamma \rangle$. 
%Here, $\mathcal{S}$ denotes the state space representing all possible environment configurations, $\mathcal{A}$ the action space of feasible robot actions, $\mathcal{P}(s_{t+1}|s_t,a_t)$ the state transition probability, $\mathcal{R}(s_t,a_t)$ the reward function providing immediate feedback, and $\gamma \in (0,1)$ the discount factor determining the importance of future rewards.

Given the limited sensing range of onboard sensors, the crowd-navigation problem is typically formulated as a partially observable Markov decision process (POMDP), defined by the tuple $\langle \mathcal{S}, \mathcal{A}, \mathcal{P}, \mathcal{R}, \gamma \rangle$. Here, $\mathcal{S}$ denotes the state space of all possible environment configurations; $\mathcal{A}$ the set of feasible robot actions; $\mathcal{P}(s_{t+1}\mid s_t, a_t)$ the state-transition probability; $\mathcal{R}(s_t, a_t)$ the reward function providing immediate feedback; and $\gamma \in (0,1)$ the discount factor governing the weighting of future rewards.

%The state space $\mathcal{S} = \{ s_0, s_1, \ldots, s_t, \ldots \}$ represents all possible environment states. 
%At time $t$, the state $s_t$ is composed of the ego-robot state $\mathbf{w}^t$ and the states of surrounding pedestrian agents $\mathbf{u}_i^t$. 
%All ego-robot information is fully observable, while the states of other agents are only partially observable. 
%The ego-robot state $\mathbf{w}^t$ includes its position $(p_x, p_y)$, velocity $(v_x, v_y)$, goal position $(g_x, g_y)$, maximum speed $v_{\max}$, heading angle $\theta$, and radius $\rho$. 
%Each pedestrian state $\mathbf{u}_i^t$ consists of the agent’s position $(p_x^i, p_y^i)$ and radius $\rho^i$.

The state space $\mathcal{S} = \{s_0, s_1, \ldots, s_t, \ldots\}$ represents all possible environment states. At time $t$, the state $s_t$ comprises the ego-robot state $\mathbf{w}^t$ and the states of surrounding pedestrian agents $\mathbf{u}_i^t$. The ego-robot state is fully observable, whereas the states of other agents are only partially observable. The ego-robot state $\mathbf{w}^t$ includes its position $(p_x, p_y)$, velocity $(v_x, v_y)$, goal position $(g_x, g_y)$, maximum speed $v_{\max}$, heading angle $\theta$, and radius $\rho$. Each pedestrian state $\mathbf{u}_i^t$ consists of the agent’s position $(p_x^i, p_y^i)$ and radius $\rho^i$.

%At the beginning of each episode, the robot starts from an initial state $s_0 \in \mathcal{S}_0$. At each timestep $t$, it selects an action $a_t \in \mathcal{A}$ according to its policy $\pi(a_t \mid s_t)$, which maps the current observable state $s_t$ to a probability distribution over feasible actions. After executing the action, the robot receives a reward $r_t$ and transitions to the next state $s_{t+1}$ following $\mathcal{P}(s_{t+1} \mid s_t, a_t)$, which represents the stochastic dynamics of the environment, including interactions with nearby humans. Each human agent also acts according to its own policy, and the joint actions of all agents govern the evolution of the overall system state. The episode terminates when the robot reaches its goal, exceeds the maximum timestep $T$, or collides with a human.

At the beginning of each episode, the robot starts from an initial state $s_0 \in \mathcal{S}_0$. At each timestep $t$, it selects an action $a_t \in \mathcal{A}$ according to its policy $\pi(a_t \mid s_t)$, which maps the current observable state $s_t$ to a probability distribution over feasible actions. After executing the action, the robot receives a reward $r_t$ and transitions to the next state $s_{t+1}$ following $\mathcal{P}(s_{t+1} \mid s_t, a_t)$, which represents the stochastic dynamics of the environment, including interactions with nearby humans. Each human agent also acts according to its own policy, and the joint actions of all agents govern the evolution of the overall system state. The episode terminates when the robot reaches its goal, exceeds the maximum timestep $T$, or collides with a human.

\begin{figure}[t]
	\begin{center}
		\includegraphics[width=0.85\linewidth]{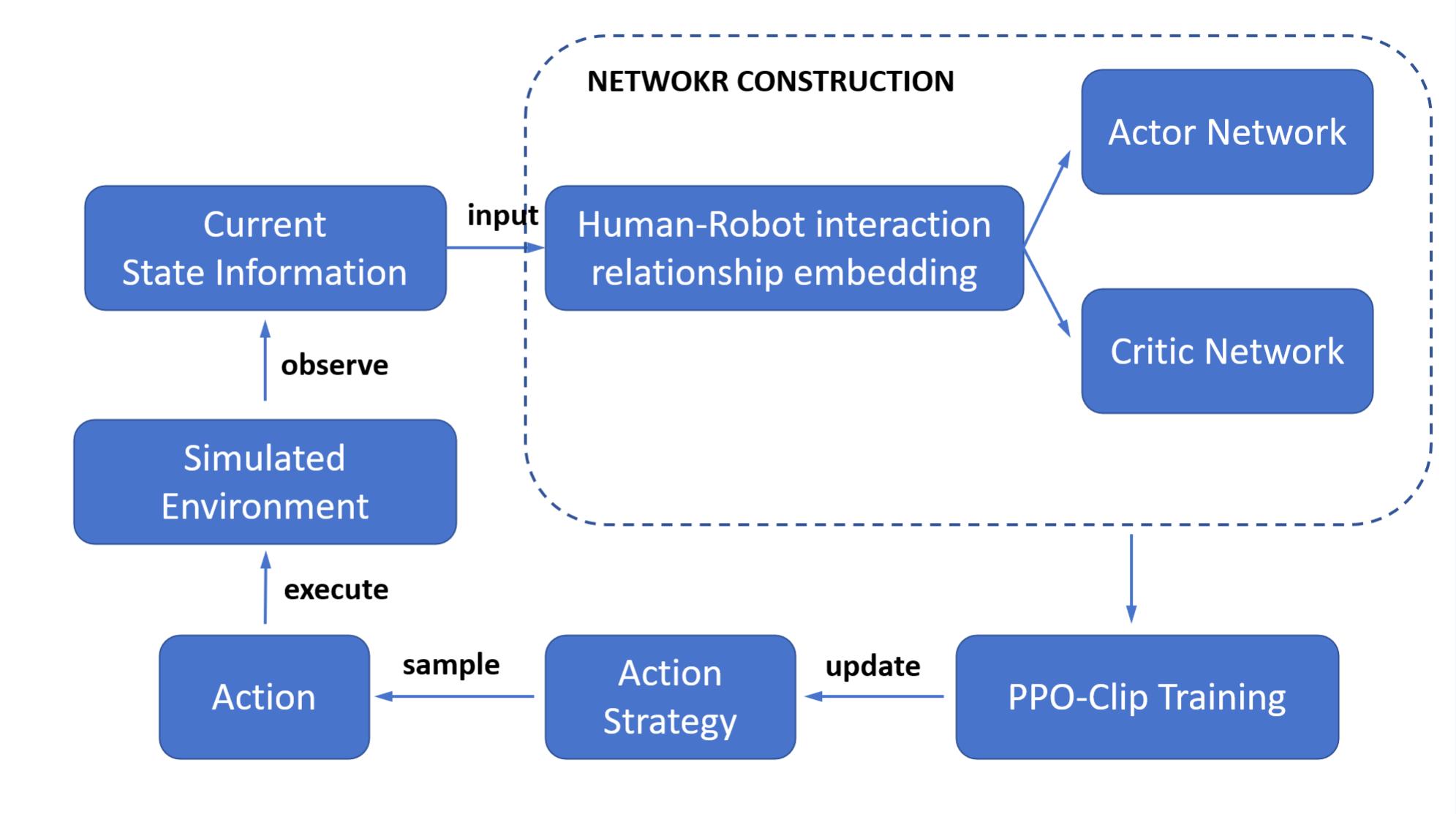}
	\end{center}
	\captionsetup{font=small}
	\caption{The schematic diagram of PPO training process for crowd navigation.}
	\label{fig:frame}
\end{figure}

Deep reinforcement learning (DRL) provides an effective framework for approximating optimal policies under partial observability by using deep neural networks for state representation and decision-making. In this study, Proximal Policy Optimisation (PPO)~\cite{schulman2017proximal} is adopted as the primary algorithm. PPO supports a continuous action space for fine-grained motion planning and is relatively insensitive to hyperparameter tuning, which facilitates reward shaping and stabilises training. The training pipeline is illustrated in Fig.~\ref{fig:frame}. The human--robot interaction embedding module mainly employs an attention-based interaction graph~\cite{liu2023intention}, whose output serves as the input to both the actor and critic networks.

PPO is a widely used policy-gradient method that enhances training stability while maintaining sample efficiency. Its objective is to maximise the expected return by updating the policy parameters~$\theta$:
\begin{equation}
	L^{PG}(\theta)=\mathbb{E}_t\!\left[\log \pi_\theta(a_t \mid s_t)\,\hat{A}_t\right],
	\label{eq:ppo_pg}
\end{equation}
where $\pi_\theta$ denotes the parameterised policy and $\hat{A}_t$ the estimated advantage at timestep~$t$. However, large policy updates can lead to instability. To address this, PPO introduces a clipped surrogate objective. Defining the probability ratio as
\begin{equation}
	r_t(\theta)=\frac{\pi_\theta(a_t \mid s_t)}{\pi_{\theta_{\text{old}}}(a_t \mid s_t)},
	\label{eq:ppo_ratio}
\end{equation}
the clipped objective becomes
{\small
	\begin{equation}
		L^{CLIP}(\theta)=\mathbb{E}_t\!\left[
		\min\!\Big(r_t(\theta)\hat{A}_t,\;
		\text{clip}\big(r_t(\theta),\,1-\epsilon,\,1+\epsilon\big)\hat{A}_t\Big)
		\right],
		\label{eq:ppo_clip}
	\end{equation}
}
where $\epsilon$ is a small constant constraining the update step to prevent destructive policy shifts. This formulation provides a favourable balance between exploration and stability, making PPO computationally efficient, robust, and widely adopted across both continuous and discrete control domains.

%We adopt the reward function formulation following, which simultaneously encourages task completion and penalizes unsafe or uncomfortable behaviors. The reward at timestep $t$ is defined as:
%
%{\small
%	\begin{equation}
%		R_t(s_t^{jn}, a_t) =
%		\begin{cases}
%			-0.25, & \text{if } d_t < 0 \quad (\text{collision penalty}),\\[4pt]
%			-0.1 + \dfrac{d_t}{2}, & \text{if } 0 \le d_t < 0.2 \quad (\text{proximity penalty}),\\[6pt]
%			1, & \text{if } \mathbf{p}_t = \mathbf{p}_g \quad (\text{goal reward}),\\[4pt]
%			0, & \text{otherwise.}
%		\end{cases}
%		\label{eq:reward_function}
%	\end{equation}
%}

We adopt the following reward formulation, which jointly encourages task completion while penalising unsafe or uncomfortable behaviours. The reward at timestep~$t$ is defined as

{\small
	\begin{equation}
		R_t(s_t^{jn}, a_t) =
		\begin{cases}
			-0.25, & \text{if } d_t < 0 \quad (\text{collision penalty}),\\[4pt]
			-0.1 + \dfrac{d_t}{2}, & \text{if } 0 \le d_t < 0.2 \quad (\text{proximity penalty}),\\[6pt]
			1, & \text{if } \mathbf{p}_t = \mathbf{p}_g \quad (\text{goal reward}),\\[4pt]
			0, & \text{otherwise}.
		\end{cases}
		\label{eq:reward_function}
	\end{equation}
}

%Here, $d_t$ denotes the minimum separation distance between the robot and surrounding humans during the interval $[t-\Delta t,\, t]$, and $\mathbf{p}_t$ and $\mathbf{p}_g$ represent the current and goal positions of the robot, respectively. 
%
%The reward components are defined as follows:
%\begin{itemize}
%	\item \textit{Collision penalty:} When a collision occurs ($d_t < 0$), a strong negative reward of $-0.25$ is assigned to discourage unsafe behaviors.
%	\item \textit{Proximity penalty:} When the robot approaches too close to a human ($0 \le d_t < 0.2$), a small negative reward proportional to distance encourages comfortable navigation.
%	\item \textit{Goal-reaching reward:} When the robot reaches its goal ($\mathbf{p}_t = \mathbf{p}_g$), it receives a positive reward of $+1$ to reinforce task completion.
%	\item \textit{Otherwise:} No reward or penalty is applied.
%\end{itemize}

Here, $d_t$ denotes the minimum separation distance between the robot and surrounding humans over the interval $[t-\Delta t,\, t]$, and $\mathbf{p}_t$ and $\mathbf{p}_g$ represent the robot’s current and goal positions, respectively. The reward components are interpreted as follows:

\begin{itemize}
	\item \textit{Collision penalty:} When a collision occurs ($d_t < 0$), a strong negative reward of $-0.25$ discourages unsafe behaviour.
	\item \textit{Proximity penalty:} When the robot gets too close to a human ($0 \le d_t < 0.2$), a mild penalty proportional to distance encourages comfortable navigation.
	\item \textit{Goal reward:} Upon reaching the goal ($\mathbf{p}_t = \mathbf{p}_g$), a reward of $+1$ reinforces successful task completion.
	\item \textit{Otherwise:} No reward or penalty is applied. (\textbf{revise})
\end{itemize}

\subsubsection{Current Commonly Used Indicators}
\label{sec:3.1}
%Common evaluation indicators in crowd navigation primarily focus on three aspects: safety, efficiency, and comfort. 
%Safety emphasizes collision avoidance and remains the foremost requirement in navigation. 
%Efficiency measures how quickly the robot completes its task, whereas comfort reflects the preservation of pedestrian comfort, typically represented by interaction distance. 
%These two objectives are often conflicting—improving efficiency may reduce comfort, while prioritizing comfort can lower efficiency. 
%Thus, ensuring safety while balancing the two remains a central challenge in crowd navigation research. 
%
%The commonly used evaluation metrics are defined as follows:

Common evaluation indicators in crowd navigation mainly focus on three aspects: safety, efficiency, and comfort. Safety, centred on collision avoidance, is the foremost requirement. Efficiency measures how quickly the robot completes its task, whereas comfort reflects the preservation of pedestrian well-being, typically characterised by interaction distance. These two objectives often conflict: improving efficiency may compromise comfort, while prioritising comfort can reduce efficiency. Consequently, ensuring safety while balancing the two remains a central challenge in crowd-navigation research.

The commonly used evaluation metrics are defined as follows:

%\begin{itemize}
%	\item \textit{Success Rate ($M_{sr}$)}: Ratio of episodes in which the robot reaches its goal without collision.
%	\item \textit{Collision Rate ($M_{cr}$)}: Ratio of episodes in which the robot collides with other agents.
%	\item \textit{Timeout Rate ($M_{tr}$)}: Ratio of episodes where the robot fails to reach the goal within the time limit (30\,s) due to the freezing robot problem.
%	\item \textit{Average Time ($M_{at}$)}: Average time (in seconds) for the robot to successfully reach its goal.
%	\item \textit{Discomfort Number ($M_{dr}$)}: Total rate of discomfort cases, defined as instances where the distance between the robot and humans is less than 0.5\,m without collision.
%	\item \textit{Minimum Distance ($M_{md}$)}: Average minimum distance between the robot and humans across all test cases.
%\end{itemize}
%
%Although existing studies employ various comprehensive metrics, few quantitatively evaluate trajectory continuity. 
%As discussed earlier, trajectory continuity is critical to both comfort and energy efficiency. 
%Moreover, trade-offs among multiple objectives remain challenging: prioritizing safety often leads to conservative and inefficient behaviors, while overemphasizing efficiency or comfort can compromise safety and overall performance.

\begin{itemize}
	\item \textit{Success Rate ($M_{sr}$)}: The proportion of episodes in which the robot reaches its goal without collision.
	\item \textit{Collision Rate ($M_{cr}$)}: The proportion of episodes in which the robot collides with other agents.
	\item \textit{Timeout Rate ($M_{tr}$)}: The proportion of episodes in which the robot fails to reach the goal within the time limit (30\,s), typically due to the freezing-robot problem.
	\item \textit{Average Time ($M_{at}$)}: The average time (seconds) required for successful goal completion.
	\item \textit{Discomfort Number ($M_{dr}$)}: The total rate of discomfort cases, defined as instances where the robot comes within 0.5\,m of a human without colliding.
	\item \textit{Minimum Distance ($M_{md}$)}: The average minimum robot--human separation across all test cases.
\end{itemize}

Although existing studies employ several holistic metrics, few provide a quantitative assessment of trajectory continuity. As discussed earlier, trajectory continuity is crucial for both comfort and energy efficiency. Moreover, balancing multiple objectives remains challenging: prioritising safety often results in overly conservative and inefficient behaviour, whereas overemphasising efficiency or comfort may compromise safety and degrade overall performance.

\subsection{The Design of a New and More Comprehensive Indicator System}

\subsubsection{Design of Trade-Offs In The Comprehensive Indicator System}

We categorise the comprehensive evaluation into five key aspects: \textit{safety}, \textit{success rate}, \textit{comfort}, \textit{trajectory quality}, and \textit{efficiency}. Among these, \textit{safety} holds the highest priority in real-world pedestrian environments, as it is essential for both pedestrian and robot protection. Although multi-scenario navigation strategies often achieve success rates above 90\%, even infrequent collisions may accumulate over long-term operation, potentially causing hardware damage, environmental disruption, or personal injury. Consequently, this study treats \textit{safety} as the foremost objective, followed by \textit{success rate}, since reaching the goal remains the fundamental requirement of navigation. Even if this entails compromises in \textit{comfort}, \textit{efficiency}, or \textit{trajectory quality}, successful task completion is the minimum criterion for effective navigation.

The remaining criteria—\textit{comfort}, \textit{trajectory quality}, and \textit{efficiency}—reflect trade-offs between performance and human experience. Prior studies have shown that \textit{comfort} is primarily influenced by interpersonal distance, while \textit{trajectory quality} affects comfort indirectly by improving motion predictability. From the perspective of pedestrian comfort, it is reasonable to sacrifice some \textit{efficiency} or \textit{trajectory smoothness} to maintain appropriate spacing. However, excessive concessions in \textit{efficiency} may lead to overly conservative or unnecessarily circuitous trajectories.

To obtain a unified quantitative assessment that reflects multiple evaluation dimensions, all normalised indicators are combined through a weighted aggregation scheme. Let
\begin{equation}
	F_{saf},\; F_{suc},\; F_{comf},\; F_{traj},\; F_{effic} \in [0,1]
\end{equation}
denote the normalised scores for \textit{safety}, \textit{success rate}, \textit{comfort}, \textit{trajectory quality}, and \textit{efficiency}, respectively. The overall performance index~$F$ is defined as
\begin{align}
	F &= w_{saf}F_{saf} + w_{suc}F_{suc} + w_{comf}F_{comf} \notag\\
	&\quad + w_{traj}F_{traj} + w_{effic}F_{effic},
\end{align}
where $w_{saf}, w_{suc}, w_{comf}, w_{traj}, w_{effic}$ are the corresponding weighting coefficients satisfying $\sum_i w_i = 1$.

This formulation provides a balanced and interpretable evaluation across heterogeneous performance criteria. Following the \textit{safety-priority} principle adopted in this study, the weighting coefficients are assigned to emphasise safety and success while preserving the influence of the remaining secondary factors. Their relative ordering is given by
\begin{equation}
	w_{saf} > w_{suc} > w_{comf} \gtrsim w_{traj} \gtrsim w_{effic}.
\end{equation}

This weighting scheme reflects our \textit{safety-priority} philosophy, ensuring that \textit{safety} and \textit{success rate} dominate the overall assessment, while \textit{comfort}, \textit{trajectory quality}, and \textit{efficiency} serve as secondary yet meaningful indicators for finer performance discrimination.

In this study, the weighting values are assigned as follows: $w_{saf} = 0.40$, $w_{suc} = 0.25$, $w_{comf} = 0.15$, $w_{traj} = 0.12$, and $w_{effic} = 0.08$.

\subsubsection{A New Metric in Crowd Navigation for Evaluating Trajectory Quality}
%In this work, we introduce a trajectory optimization metric designed to evaluate and enhance 
%the geometric smoothness of robot motion. 
%Specifically, the trajectory quality is assessed by computing the curvature at consecutive trajectory points 
%and applying a threshold-based criterion. 
%When the curvature difference between adjacent segments exceeds this threshold, 
%the trajectory is considered to exhibit a $C^2$ discontinuity, indicating a loss of geometric continuity. 
%To quantitatively characterize this property, a curvature difference metric is formulated based on 
%discrete trajectory points. 
%Assume that a trajectory consists of $N$ consecutive points, denoted as:

In this work, we introduce a trajectory-optimization metric designed to evaluate and enhance the geometric smoothness of robot motion. Trajectory quality is assessed by computing the curvature at consecutive trajectory points and applying a threshold-based criterion. When the curvature difference between adjacent segments exceeds this threshold, the trajectory is regarded as exhibiting a $C^{2}$ discontinuity, indicating a loss of geometric continuity. To quantify this property, we formulate a curvature-difference metric based on discrete trajectory points.

Assume that a trajectory consists of $N$ consecutive points, denoted as:

\begin{equation}
	P_i = (x_i, y_i), \quad i = 1, 2, \ldots, N.
\end{equation}

In a two-dimensional plane, the local curvature determined by three consecutive points 
\(P_i, P_{i+1}, P_{i+2}\) is given by:
\begin{equation}
	\kappa_i =
	\frac{2 |(x_{i+1}-x_i)(y_{i+2}-y_i) - (y_{i+1}-y_i)(x_{i+2}-x_i)|}
	{\sqrt{d_{i,i+1}^2\, d_{i+1,i+2}^2\, d_{i,i+2}^2}},
	\label{eq:curvature_short}
\end{equation}
where \( d_{m,n} = \sqrt{(x_n - x_m)^2 + (y_n - y_m)^2} \) 
is the Euclidean distance between two points.

Eq.~\ref{eq:curvature_short} defines the discrete curvature, which corresponds to the reciprocal of the radius of the circumcircle through the three points, thereby describing the local bending of the trajectory.

Given four consecutive points 
\(P_i, P_{i+1}, P_{i+2}, P_{i+3}\), 
the curvatures of two adjacent segments are computed as:
\begin{align}
	\kappa_1 &= f(P_i, P_{i+1}, P_{i+2}), \nonumber \\
	\kappa_2 &= f(P_{i+1}, P_{i+2}, P_{i+3}),
	\label{eq:two point curvature}
\end{align}
where \( f(\cdot) \) denotes the curvature function defined in Eq.~\ref{eq:curvature_short}.

The curvature difference between the two neighbouring segments is then
\begin{equation}
	\Delta \kappa_i = |\kappa_2 - \kappa_1|.
	\label{eq:delta_curvature}
\end{equation}

To evaluate local smoothness, a curvature-difference threshold \(\tau\) is introduced.  
If the curvature variation is smaller than \(\tau\), the segment is regarded as smooth and assigned a value of~1; otherwise, it is assigned~0:
\begin{equation}
	C_i =
	\begin{cases}
		1, & \text{if } |\kappa_2 - \kappa_1| < \tau, \\
		0, & \text{otherwise.}
	\end{cases}
	\label{eq:count}
\end{equation}

Traversing all four-point groups along the trajectory, the overall smoothness measure \(M_{cdr}\) is computed as:
\begin{equation}
	M_{cdr}= \sum_{i=1}^{N-3} C_i.
	\label{eq:smoothness_sum}
\end{equation}

Expanding Eqs.~\ref{eq:curvature_short}--\ref{eq:smoothness_sum}, the complete expression for the trajectory-smoothness metric is:
\begin{equation}
	\begin{aligned}
		M_{cdr} = \sum_{i=1}^{N-3}
		\mathbf{1} \Big(
		& \big|
		f(P_{i+1}, P_{i+2}, P_{i+3})
		\\
		& {} -
		f(P_i, P_{i+1}, P_{i+2})
		\big| < \tau
		\Big),
	\end{aligned}
	\label{eq:final_smoothness_split}
\end{equation}
where \(\mathbf{1}(\cdot)\) is the indicator function, equal to~1 if the condition holds and~0 otherwise.

\subsubsection{Specific Evaluation Methodology for Each Aspect}

\paragraph{Security Assessment.} In our evaluation framework, safety is assigned the highest priority. To reflect this, the safety score $F_{saf}$ is designed to exhibit heightened sensitivity to small variations when the collision rate is low, thereby enabling fine discrimination within the safe region. A smooth threshold function is adopted to provide a continuous and differentiable mapping from the collision rate $M_{cr}$ to the corresponding safety score $F_{saf}$. The function is defined as
\begin{equation}
	F_{saf}(M_{cr}) = \frac{1}{1 + \left( \dfrac{M_{cr}}{\tau_S} \right)^{\beta}},
	\label{eq:safety_smooth}
\end{equation}
where $\tau_S$ denotes the safety threshold and $\beta$ controls the steepness of the curve. In this paper, $\tau_S$ is set to 0.05 in the low-density scenario and 0.1 in the high-density scenario, while the shaping parameter is fixed at $\beta = 4$. This configuration ensures that $F_{saf}$ decreases smoothly yet perceptibly as the collision rate increases.

This function satisfies the following properties:
\begin{equation}
	F_{saf}(0) = 1, \quad 
	F_{saf}(\tau_S) = 0.5, \quad 
	\text{and} \quad 
	F_{saf}(M_{cr} \gg \tau_S) \to 0.
	\label{eq:safty_example}
\end{equation}

%In our evaluation framework, safety is assigned the highest priority. 
%To reflect this, the safety score $F_{saf}$ is designed to exhibit higher sensitivity to small variations when the collision rate is low, ensuring a fine distinction in the safe region. 
%A smooth-threshold function is adopted to provide a continuous and differentiable mapping between the collision rate $M_{cr}$ and the corresponding safety score $F_{saf}$. 
%The function is defined as
%\begin{equation}
%	F_{saf}(M_{cr}) = \frac{1}{1 + \left( \dfrac{M_{cr}}{\tau_S} \right)^{\beta}},
%	\label{eq:safety_smooth}
%\end{equation}
%where $\tau_S$ denotes the safety threshold, 
%and $\beta$ controls the steepness of the curve. 
%In this paper, $\tau_S$ is set to 0.05 in the low-density scenario and 0.1 in the high-density scenario, while the shaping parameter is fixed at $\beta = 4$. This configuration ensures that $F_{saf}$ decreases smoothly yet noticeably as the collision rate increases.
%
%This function satisfies the following properties:
%\begin{equation}
%	F_{saf}(0) = 1, \quad F_{saf}(\tau_S) = 0.5, \quad \text{and} \quad F_{saf}(cr \gg \tau_S) \to 0.
%	\label{eq:safty_example}
%\end{equation}

\paragraph{Success Assessment.} The success rate can be directly adopted as its own assessment metric:
\begin{equation}
	F_{suc}(M_{sr}) = M_{sr}.
	\label{eq:success assessment}
\end{equation}

\paragraph{Comfort Assessment.}
Comfort is primarily influenced by two indicators: the discomfort frequency $M_{dr}$ and the minimum interpersonal distance $M_{md}$. Lower values of $M_{dr}$ and higher values of $M_{md}$ correspond to improved comfort performance. Given the desired upper bound $\tau^{\min}_{md}$ for $M_{md}$, linear scaling and truncation are applied to compute each sub-term, and the overall comfort score is then obtained by a weighted aggregation. The formulation is as follows:
\begin{align}
	F^{dn}_{comf} &= (1 - M_{dr})^{\gamma}, \quad \gamma > 0, \notag\\[4pt]
	F^{md}_{comf} &= \mathrm{clip}\!\left(\frac{M_{md}}{\tau^{\min}_{md}},\, 0,\, 1\right),
	\label{eq:comfort_ca_cb}
\end{align}
where $F^{dn}_{comf}$ denotes the term associated with discomfort frequency, and $F^{md}_{comf}$ the term associated with minimum distance. The exponent $\gamma$ controls the sensitivity to $M_{dr}$; when $\gamma > 1$, the function becomes more sensitive to higher discomfort ratios. To enhance the discrimination of subtle differences, $\gamma$ is set to 10 in this study.

The overall comfort score $F_{comf}$ is computed as:
\begin{equation}
	F_{comf} = \lambda\,F^{dn}_{comf} + (1-\lambda)\,F^{md}_{comf},
	\qquad \lambda \in [0,1],
	\label{eq:comfort_final}
\end{equation}
where $\lambda$ controls the relative importance between discomfort frequency and proximity, and is set to 0.5 in this work.

\paragraph{Trajectory Assessment.}

A simple and monotonic power-law inverse mapping is adopted to convert the \textit{curvature discontinuity ratio} ($M_{cdr} \in [0,1]$), defined in Section~xx, into a normalised \textit{trajectory quality score} ($F_{traj} \in [0,1]$):
\begin{equation}
	F_{traj}(M_{cdr}) = (1 - M_{cdr})^{\gamma}, \qquad \gamma > 0,
	\label{eq:trajectory_quality_power}
\end{equation}
This formulation provides an interpretable and tunable mapping from curvature discontinuity to trajectory quality, enabling seamless integration into the multi-objective evaluation framework. The parameter~$\gamma$ is set identically to that used in $F^{dn}_{\text{comf}}$.

\paragraph{Efficiency Assessment.}
Navigation efficiency is evaluated by comparing the \textit{Average Time} ($M_{at}$) with the theoretical best-case time, or \textit{Optimal Time} ($T^{*}$), which represents the straight-line travel time at maximum speed in the absence of obstacles. The efficiency score is computed as:
\begin{equation}
	F_{effic}(M_{at}) = \min\!\left(1,\, \frac{T^{*}}{M_{at}}\right).
	\label{eq:efficiency_power}
\end{equation}

%A simple and monotonic power-law inverse mapping is then applied to convert the \textit{curvature discontinuity ratio} ($M_{cdr} \in [0,1]$), defined in Section~xx, into a normalized \textit{trajectory quality score} ($F_{traj} \in [0,1]$).
%\begin{equation}
%	F_{traj}(M_{cdr}) = (1 - M_{cdr})^{\gamma}, \quad \gamma > 0,
%	\label{eq:trajectory_quality_power}
%\end{equation}
%This formulation offers an interpretable and tunable mapping from the curvature discontinuity ratio to trajectory quality, facilitating its integration into multi-objective evaluation frameworks. The parameter $\gamma$ follows the same setting as that used in $F^{dn}_{\text{comf}}$.
%
%\paragraph{Efficiency Assessment.}
%To evaluate navigation efficiency, the \textit{Average Time} ($M_{at}$) is compared with the theoretical fastest reachable time, or \textit{Optimal Time} ($T^{*}$), which represents the time required to reach the destination in a straight line at maximum speed without obstacles. 
%The specific formula is as follows.
%\begin{equation}
%	F_{effic}(M_{at}) = \min\!\left(1, \frac{T^{*}}{M_{at}}\right),
%	\label{eq:efficiency_power}
%\end{equation}

\subsection{DRL-based trajectory optimization}
\label{sec:4.2}
\subsubsection{Continuous Action Space}
\label{sec:4.2.1}
%In this study, a continuous action space is employed to investigate the impact and execution differences between fine-grained and coarse-grained actions on trajectory generation. 
%On one hand, finer actions enable more precise exploration of advantageous behaviors, and when applied consistently, their cumulative effect can significantly enhance overall performance. 
%On the other hand, coarse-grained actions are discretized, limiting the optimization of trajectory quality and the smoothness of control. The robot is modeled with holonomic kinematics. At each timestep $t$, their action consists of a desired velocity vector
%\begin{equation}
%	\mathbf{a}_t = [v_x,\, v_y],
%	\label{eq:action_space}
%\end{equation}
%where $v_x$ and $v_y$ denote the velocity components along the $x$- and $y$-axes, respectively. The robot’s action space is continuous, with a maximum speed of $1$ m/s. In continuous control PPO, the policy outputs the mean and standard deviation of a Gaussian distribution over actions. 
%Actions are sampled as:
%\begin{equation}
%	a_t^{\text{raw}} = 
%	\mu_{\theta}(s_t) + \sigma_{\theta}(s_t) \odot \epsilon_t,
%	\quad \epsilon_t \sim \mathcal{N}(0, I),
%\end{equation}
%and optionally squashed by $\tanh$ to fit the action bounds.
%The corresponding log-probabilities are used to compute the PPO 
%objective ratio for stable policy optimization.

In this study, a continuous action space is employed to examine the impact and behavioural differences between fine-grained and coarse-grained actions on trajectory generation. Finer actions allow more precise exploration of advantageous behaviours and, when applied consistently, their cumulative effect can substantially enhance overall performance. In contrast, coarse-grained discretised actions restrict the optimisation of trajectory quality and control smoothness.

The robot is modelled with holonomic kinematics. At each timestep~$t$, its action is defined as a desired velocity vector:
\begin{equation}
	\mathbf{a}_t = [v_x,\, v_y],
	\label{eq:action_space}
\end{equation}
where $v_x$ and $v_y$ denote the velocity components along the $x$- and $y$-axes, respectively. The action space is continuous, with a maximum speed of $1\,\text{m/s}$. In continuous-control PPO, the policy outputs the mean and standard deviation of a Gaussian distribution over actions, from which actions are sampled as:
\begin{equation}
	a_t^{\text{raw}} =
	\mu_{\theta}(s_t) + \sigma_{\theta}(s_t) \odot \epsilon_t,
	\qquad \epsilon_t \sim \mathcal{N}(0, I),
\end{equation}
optionally followed by a $\tanh$ squashing transformation to enforce action bounds. The corresponding log-probabilities are then used to compute the PPO objective ratio for stable policy optimisation.

\begin{figure}[t]
	\begin{center}
		\includegraphics[width=0.85\linewidth]{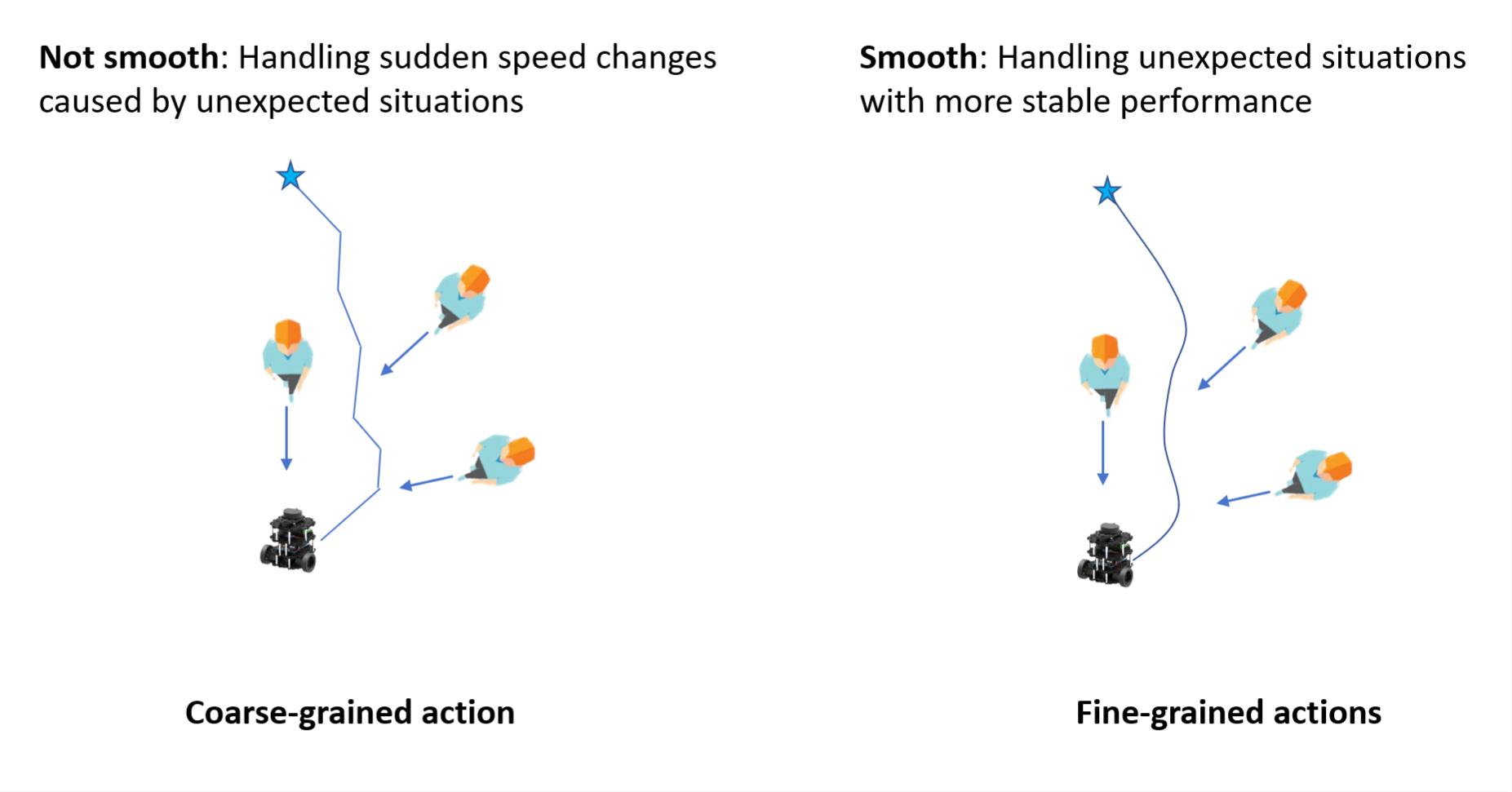}
	\end{center}
	\captionsetup{font=small}
	\caption{The difference between the impact of continuous action space and discrete action space on trajectory.}
	\label{fig:action_space}
\end{figure}

\subsubsection{Reward Shaping for Improving Trajectory Quality}
\label{sec:4.2.2}
%To improve motion smoothness and suppress abrupt directional changes, a curvature-based smoothness reward term is incorporated into the reinforcement learning framework. 
%At each step, the curvature variation $\Delta \kappa$ between consecutive trajectory segments is used to evaluate the local smoothness of motion. 
%A soft exponential mapping is employed to convert the curvature variation into a penalty term:
%\begin{equation}
%	r_{\text{curv}} =
%	\begin{cases}
%		\lambda \left(1 - e^{-|\Delta \kappa|}\right), & \text{if } 1 - e^{-|\Delta \kappa|} > \tau_c, \\[4pt]
%		0, & \text{otherwise},
%	\end{cases}
%	\label{eq:curv_penalty_final}
%\end{equation}
%where $\lambda > 0$ is a scaling coefficient, and $\tau_c$ denotes the curvature change threshold (set to $0.5$ in our implementation). 
%This design softly penalizes sharp curvature changes while ignoring small natural steering variations.
%
%The final reward at each time step is formulated as:
%\begin{equation}
%	R_t = R_t^{\text{base}} - w_{\text{smooth}} \, r_{\text{curv}},
%	\label{eq:curv_total_reward_final}
%\end{equation}
%where $R_t^{\text{base}}$ is the base reward from the environment, and $w_{\text{smooth}} = 0.2$ adjusts the influence of the smoothness term. 
%
%This formulation effectively promotes globally smoother and more human-like trajectories. 
%The exponential transformation guarantees a continuous gradient, ensuring stable optimization, while the thresholding mechanism focuses the penalty on perceptually significant curvature fluctuations.

To improve motion smoothness and suppress abrupt directional changes, a curvature-based smoothness reward term is incorporated into the reinforcement learning framework. At each timestep, the curvature variation $\Delta \kappa$ between consecutive trajectory segments is used to assess local motion smoothness. A soft exponential mapping is employed to convert this variation into a penalty term:
\begin{equation}
	r_{\text{curv}} =
	\begin{cases}
		\lambda \left(1 - e^{-|\Delta \kappa|}\right), 
		& \text{if } 1 - e^{-|\Delta \kappa|} > \tau_c, \\[4pt]
		0, & \text{otherwise},
	\end{cases}
	\label{eq:curv_penalty_final}
\end{equation}
where $\lambda > 0$ is a scaling coefficient and $\tau_c$ is the curvature-change threshold (set to $0.5$ in our implementation). This formulation softly penalises sharp curvature deviations while ignoring small, natural steering adjustments.

The final reward at each timestep is defined as:
\begin{equation}
	R_t = R_t^{\text{base}} - w_{\text{smooth}}\, r_{\text{curv}},
	\label{eq:curv_total_reward_final}
\end{equation}
where $R_t^{\text{base}}$ is the base environmental reward and $w_{\text{smooth}} = 0.2$ controls the influence of the smoothness term.

This design effectively encourages globally smoother and more human-like trajectories. The exponential transformation ensures a continuous gradient for stable optimisation, while the thresholding mechanism targets penalisation on perceptually meaningful curvature fluctuations.

\section{EXPERIMENTS}
\label{sec:4}
\subsection{Experiments Setup}
\label{sec:5.1}
\subsubsection{Simulation Environment}
\label{sec:5.1.1}
%The 2D simulation environment is built upon Crowdsim~\cite{chen2019crowd}, \cite{liu2023intention}. 
%Two scenarios are used for training and testing. 
%In the low-density scene, five pedestrian agents are randomly distributed along a circle with a radius of 4\,m centered in the environment. 
%In the high-density scene, twenty pedestrians are randomly positioned along a circle with a radius of 6\,m, as shown in Fig.~\ref{fig:2d simulation}. 
%The robot is modeled with holonomic kinematics and a circular sensing range of 5\,m. 
%To maintain continuous pedestrian flow, each human receives a new random goal upon reaching the previous one and may occasionally switch to another random target. 
%Human motion follows the ORCA algorithm. 
%In this ``invisible robot'' setting, pedestrians are assumed unable to see the robot, effectively preventing the robot from learning overly aggressive behaviors that force humans to yield. 
%Agents are further assumed to instantaneously achieve and maintain their desired velocities for the next $\Delta t$ seconds.

\begin{figure}[t]
	\hspace{-4.5mm}
	\subfloat[Low-Density Scenario]
	{
		\includegraphics[width=0.23\textwidth]{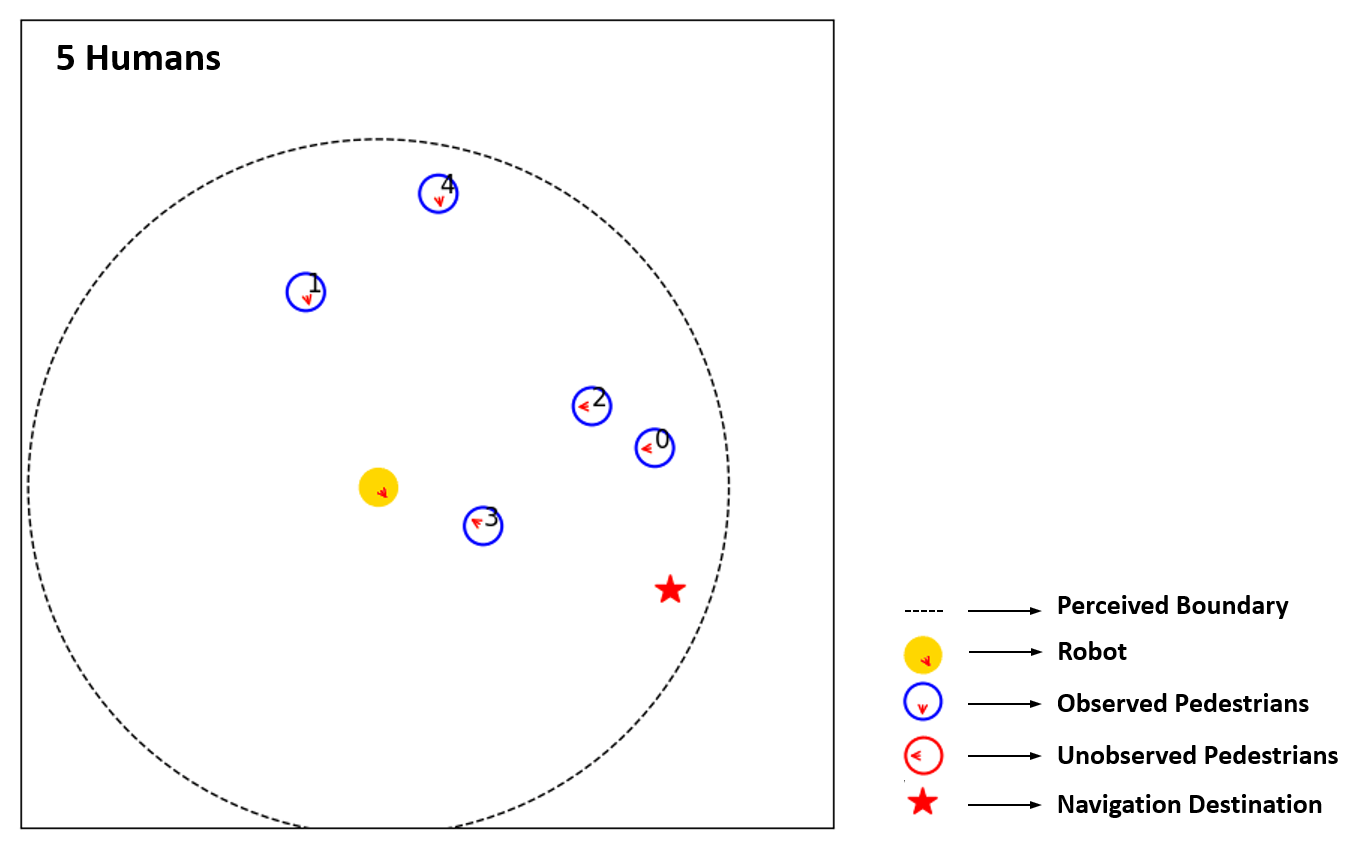}
		%\label{fig:9.1}
	}
	\subfloat[High-Density Scenario]
	{
		\includegraphics[width=0.23\textwidth]{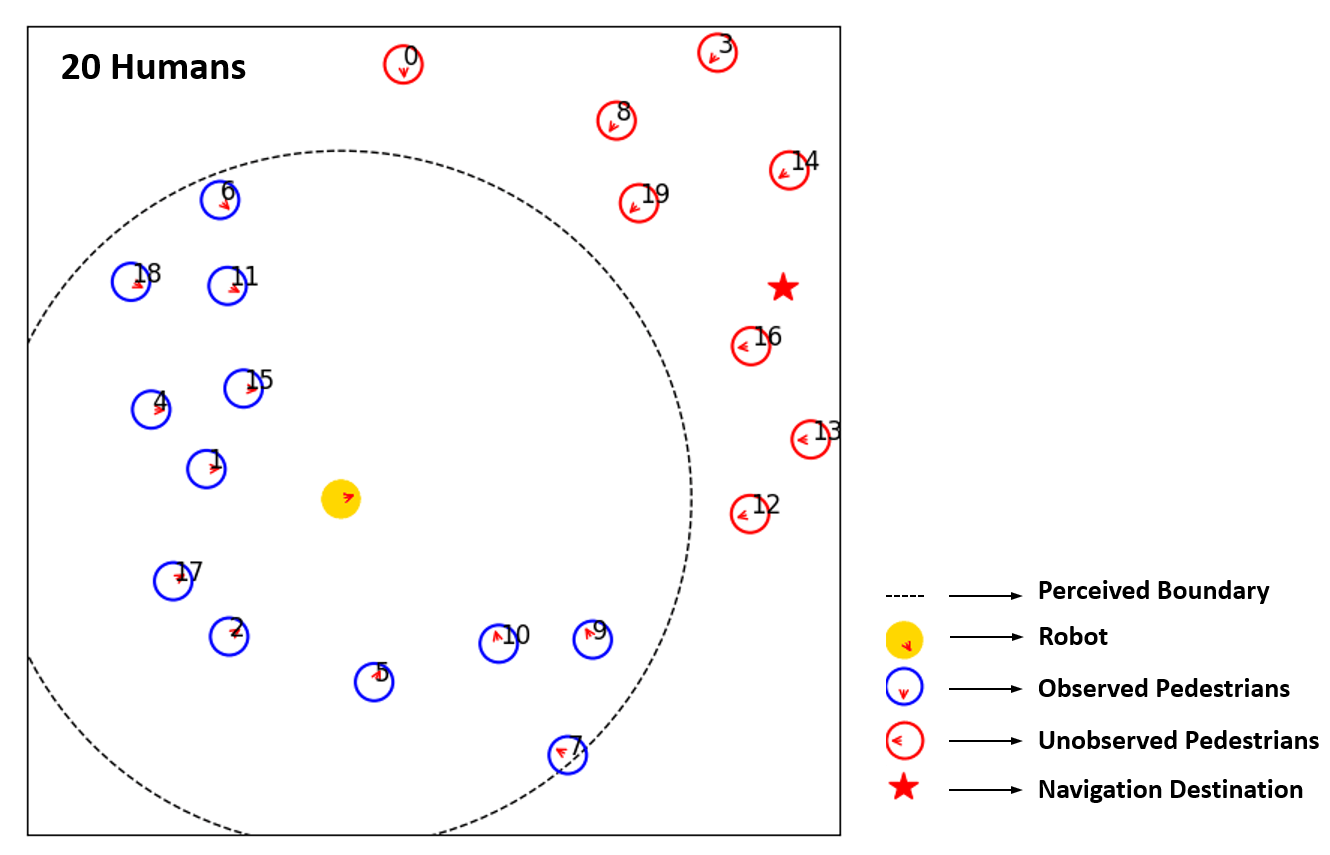}
		%\label{fig:9.2}
	}
	\captionsetup{font=small}
	\caption{2D Simulation Environment.}
	\label{fig:2d simulation}
\end{figure}

The 2D simulation environment is built upon Crowdsim~\cite{chen2019crowd, liu2023intention}. Two scenarios are used for training and testing. In the low-density setting, five pedestrian agents are randomly placed along a circle of radius 4\,m centred in the environment. In the high-density setting, twenty pedestrians are distributed along a circle of radius 6\,m, as shown in Fig.~\ref{fig:2d simulation}. The robot is modelled with holonomic kinematics and equipped with a circular sensing range of 5\,m. To maintain continuous pedestrian flow, each human agent receives a new random goal upon reaching its previous one and may occasionally switch to another randomly assigned target. Human motion is governed by the ORCA algorithm. In this ``invisible-robot'' configuration, pedestrians are assumed not to perceive the robot, thereby preventing the learning of overly aggressive behaviours in which the robot inappropriately forces humans to yield. All agents are further assumed to instantaneously achieve and maintain their desired velocities over the next $\Delta t$ seconds.

\subsubsection{Baselines and Ablation Models}
\label{sec:5.1.2}

%We adopt two widely used model-based methods, ORCA and SFM, as baseline algorithms. 
%Additionally,  \cite{liu2023intention} with the general reward section is selected as another ablation model to further validate the contribution of the proposed reward, referred to as intentionGRU in this article. 
%The method obtained by adding the proposed reward term to the IntentionGRU  with the general reward section is referred to as IntentionGRU\_Traj.

We adopt two widely used model-based methods, ORCA and SFM, as baseline algorithms. In addition, the model in~\cite{liu2023intention} using the general reward formulation is included as an ablation benchmark to validate the contribution of the proposed reward; this variant is referred to as \textit{IntentionGRU} in this paper. The method obtained by augmenting IntentionGRU with the proposed trajectory-based reward term is denoted as \textit{IntentionGRU\_Traj}.

\subsubsection{Training Process}
\label{sec:5.1.3}
%We conducted training and testing on a host computer equipped with an Intel(R) Core(TM) i7-12700 CPU and an NVIDIA RTX4080. We implemented the policy in PyTorch\cite{paszke2017automatic}. The learning rate was set to $4\times10^{-5}$, and the RMSProp optimizer was employed with parameters $\alpha = 0.99$ and $\epsilon = 1\times10^{-5}$. The discount factor was $\gamma = 0.99$, and gradient clipping was applied with a maximum norm of 0.5. Each training iteration consisted of 30 forward steps, 5 PPO epochs, a clipping parameter of 0.2, a value loss coefficient of 0.5, and an entropy coefficient of 0. Sixteen parallel environments were used with two mini-batches per update, and the total number of environment steps was $2.0\times10^{7}$. Generalized Advantage Estimation (GAE) was adopted with $\lambda = 0.95$ to enhance stability during policy updates. 
Training and testing were conducted on a host machine equipped with an Intel\,(R) Core\,(TM) i7-12700 CPU and an NVIDIA RTX\,4080 GPU. The policy was implemented in PyTorch~\cite{paszke2017automatic}. The learning rate was set to $4\times10^{-5}$, and the RMSProp optimiser was used with parameters $\alpha = 0.99$ and $\epsilon = 1\times10^{-5}$. The discount factor was $\gamma = 0.99$, and gradient clipping was applied with a maximum norm of 0.5. Each training iteration comprised 30 forward steps, 5 PPO epochs, a clipping parameter of 0.2, a value-loss coefficient of 0.5, and an entropy coefficient of 0. Additionally, sixteen parallel environments were employed with two mini-batches per update, and the total number of environment steps was $2.0\times10^{7}$. Generalised Advantage Estimation (GAE) with $\lambda = 0.95$ was adopted to improve stability during policy updates.

\subsection{Evaluation and Analysis}
\label{sec:5.2}
%In this section, we compare and evaluate the effectiveness of the proposed reward components, employing the multi-objective priority evaluation method introduced in this paper to intuitively illustrate the advantages and disadvantages of each aspect. To provide a more comprehensive evaluation, the DRL-based method was trained under both low-density and high-density scenarios. The resulting models were then extensively tested in both settings to assess their generalization performance across training and cross-density conditions. The test results are summarized in Table \ref{table:1} and Table \ref{table:2}, while the specific testing procedures are detailed in Table \ref{table:1}.

In this section, we compare and evaluate the effectiveness of the proposed reward components using the multi-objective priority evaluation method introduced earlier. This allows the advantages and limitations of each aspect to be illustrated clearly and intuitively. To provide a comprehensive assessment, the DRL-based method was trained in both low-density and high-density scenarios, and the resulting models were extensively tested in both settings to examine their generalisation across training and cross-density conditions. The test results are presented in Table~\ref{table:1} and Table~\ref{table:2}, with the corresponding testing procedures summarised in Table~\ref{table:1}.

\subsubsection{Quantitative Evaluation in Low-Density Scenarios}
\label{sec:5.2.1}

Table~\ref{table:1} presents the performance of models trained under low-density conditions. In terms of overall performance, \textit{IntentionGRU\_Traj} achieves the best results, whereas \textit{IntentionGRU} performs worse than ORCA and SFM. This is primarily due to the weaker safety performance of IntentionGRU, despite its superior trajectory quality and efficiency compared with the non-learning baselines. Incorporating the proposed trajectory-based reward markedly improves IntentionGRU, yielding a 54.5\% reduction in collisions and a 4.4\% increase in success rate, while maintaining high trajectory quality and efficiency with only minimal compromise. As a result, its overall performance surpasses that of both ORCA and SFM. When trained in high-density environments, both IntentionGRU and IntentionGRU\_Traj further improve in overall performance when evaluated in low-density conditions, indicating that high-density training facilitates effective transfer and generalisation to sparser scenarios.

\subsubsection{Quantitative Evaluation in High-Density Scenarios}
\label{sec:5.2.2}
Table~\ref{table:2} shows that models trained in high-density environments—particularly \textit{IntentionGRU\_Traj}—substantially outperform ORCA and SFM under high-density testing. IntentionGRU\_Traj attains the strongest results overall, with a 3.7\% improvement in success rate, a 24.0\% reduction in collisions, a 69.9\% reduction in trajectory discontinuities, and a 22.0\% decrease in average navigation time. When low-density models are evaluated in high-density settings, performance decreases considerably; however, IntentionGRU\_Traj consistently outperforms IntentionGRU. These findings confirm that training in high-density environments leads to stronger generalisation across varying density conditions.

\begin{table*}[!t]
	\caption{QUANTITATIVE TESTING IN LOW-DENSITY SCENARIOS }
	\centering
	\resizebox{0.95\textwidth}{!}{
		\normalsize
		\begin{tabular}{cccccccccccccc} 
			
			\toprule
			Training Scenario&Method  &$\textbf{Comprehensive}\uparrow$ &$F_{saf}\uparrow$&$F_{suc}\uparrow$&$F_{comf}\uparrow$&$F_{traj}\uparrow$ &$F_{effic}\uparrow$&$M_{sr}$($\%$)$\uparrow$&$M_{cr}$($\%$)$\downarrow$/$M_{tr}$($\%$)$\downarrow$&$M_{dr}$($\%$)$\downarrow$&$M_{md}$(m)$\uparrow$&$M_{cdr}$($\%$)$\downarrow$&$M_{at}$(s)$\downarrow$ \\
			\midrule
			\multirow{2}{*}{No Trained}&ORCA      & 0.790 & 0.642& 0.957& $\textbf{0.906}$ & 0.837 & 0.585 & 95.7 & 4.3/0.0     & 1.196 & $\textbf{0.47}$ & 1.765 & 13.686\\
			&SFM         & 0.791& $\textbf{0.993}$& 0.598& 0.877 & 0.776 & 0.251 & 59.8 & $\textbf{1.4}$/38.8      & $\textbf{0.796}$ & 0.415 & 2.5 & 31.88\\
			
			\midrule
			\multirow{2}{*}{Low-Density Scenario}&IntentionGRU        & 0.624 & 0.307& 0.929& 0.672 & $\textbf{0.979}$ & 0.631 & 92.9 & 6.6/0.5      & 5.626 & 0.392 & $\textbf{0.208}$ & 12.674  \\
			&IntentionGRU\_Traj(Ours)     	 & 0.874& 0.885& $\textbf{0.970}$& 0.784 & 0.952 & 0.575 & 97.0   & 3.0/0.0   &2.971 & 0.414 &0.486 & 13.919 \\
			\midrule
			\multirow{2}{*}{High-Density Scenario}&IntentionGRU         & 0.820& 0.773& 0.963& 0.748 & 0.916 & 0.595 & 96.3 & 3.7/0.0    & 3.945 & 0.414  & 5.061 & 13.45 \\
			&IntentionGRU\_Traj(Ours)      	& $\textbf{0.897}$& 0.95& $\textbf{0.976}$& 0.686 & 0.961 & $\textbf{0.693}$ & $\textbf{97.6}$   & 2.4/0.0   & 5.661 & 0.407 & 0.397 & $\textbf{11.544}$\\
			
			\bottomrule
		\end{tabular}
	}
	\begin{tablenotes} %���Ӵ˴�
		\footnotesize
		\item Bold values highlight the best-performing result for each metric across all methods within the same environment configuration. \textbf{Note}: For a more rigorous statistical evaluation, we utilized 10 different random seeds, each corresponding to a unique set of 500 random scenario configurations to for fair comparison. When the algorithm is tested using the same random seeds, the 500 generated scenarios remain identical. The final results, averaged from 10 sets of test values, are recorded in the Table \ref{table:1} and Table \ref{table:2}.
	\end{tablenotes}
	\label{table:1}
\end{table*}

\begin{table*}[!t]
	\caption{QUANTITATIVE TESTING IN HIGH-DENSITY SCENARIOS }
	\centering
	\resizebox{0.95\textwidth}{!}{
		\normalsize
		\begin{tabular}{cccccccccccccc} 
			
			\toprule
			Training Scenario&Method  &$\textbf{Comprehensive}\uparrow$ &$F_{saf}\uparrow$&$F_{suc}\uparrow$&$F_{comf}\uparrow$&$F_{traj}\uparrow$ &$F_{effic}\uparrow$&$M_{sr}$($\%$)$\uparrow$&$M_{cr}$($\%$)$\downarrow$/$M_{tr}$($\%$)$\downarrow$&$M_{dr}$($\%$)$\downarrow$&$M_{md}$(m)$\uparrow$&$M_{cdr}$($\%$)$\downarrow$&$M_{at}$(s)$\downarrow$ \\
			\midrule
			\multirow{2}{*}{No Trained}&ORCA      & 0.453& 0.043& 0.744& $\textbf{0.899}$ & 0.73 & 0.341 & 74.4 & 21.8/3.8    & $\textbf{2.223}$ & $\textbf{0.505}$ & 3.094 & 23.44  \\
			&SFM     & 0.303& 0.119& 0.144& 0.774 & 0.689 & 0.258 & 14.4 & 16.5/69.1    & 3.611  &0.428  & 3.659 & 31.041 \\
			
			\midrule
			\multirow{2}{*}{Low-Density Scenario}&IntentionGRU & 0.374& 0.010& 0.680& 0.541 & 0.697 & 0.435 & 68.0 & 31.2/0.8   & 11.449 & 0.393 & 3.543 & 18.395 \\
			&IntentionGRU\_Traj(Ours)    & 0.415& 0.026& 0.705& 0.637 & 0.853 & 0.381 & 70.5 & $24.7/4.8$     &7.545 & 0.409 & 1.572 & 21.004  \\
			\midrule
			\multirow{2}{*}{High-Density Scenario}&IntentionGRU & 0.547& 0.264& 0.870& 0.640 & 0.753 & 0.465 & 87.0 & 12.9/0.1   & 7.646 & 0.414  & 2.793 & 17.193 \\
			&IntentionGRU\_Traj(Ours)    & $\textbf{0.681}$& $\textbf{0.516}$& $\textbf{0.902}$& 0.606 & $\textbf{0.919}$ & $\textbf{0.596}$ & $\textbf{90.2}$ & $\textbf{9.8}$/0.0      & 8.981 & 0.41 & $\textbf{0.842}$ & $\textbf{13.418}$ \\
			
			\bottomrule
		\end{tabular}
	}
	\begin{tablenotes} %���Ӵ˴�
		\footnotesize
		\item Bold values highlight the best-performing result for each metric across all methods within the same environment configuration. 
	\end{tablenotes}
	\label{table:2}
\end{table*}

\subsubsection{Qualitative Evaluation}
\label{sec:5.2.4}
%As shown in Fig.~\ref{fig:learning_curve}, IntentionGRU exhibits unstable convergence during training in low-density scenarios, 
%whereas IntentionGRU\_Traj demonstrates relatively stable convergence in both low- and high-density settings. 
%This indicates that introducing a trajectory penalty term enhances the training stability of the model. 
%Meanwhile, A qualitative experimental analysis was conducted on ORCA, SFM, IntentionGRU, and IntentionGRU\_Traj 
%using the same high-density scene configuration, including identical start and goal positions and pedestrian motion trajectories. 
%The trajectory comparison results are illustrated in Fig. \ref{fig:traj_comp}. 
%It can be observed that SFM experiences a timeout, while ORCA produces a relatively inefficient and tortuous path. 
%IntentionGRU generates a generally smooth trajectory but exhibits abrupt directional changes near the end due to emergency obstacle avoidance. 
%In contrast, IntentionGRU\_Traj achieves a smoother and more efficient trajectory, clearly demonstrating its superior overall performance compared with other methods. 

\begin{figure}[t]
	\begin{center}
		\includegraphics[width=0.85\linewidth]{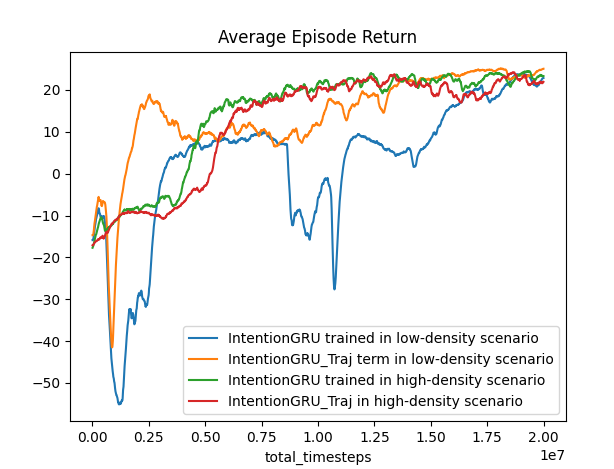}
	\end{center}
	\captionsetup{font=small}
	\caption{Convergence curve during the training process.}
	\label{fig:learning_curve}
\end{figure}

As shown in Fig.~\ref{fig:learning_curve}, \textit{IntentionGRU} exhibits unstable convergence during training in low-density scenarios, whereas \textit{IntentionGRU\_Traj} demonstrates comparatively stable convergence in both low- and high-density settings. This indicates that introducing the trajectory-penalty term enhances the stability of the training process.

\begin{figure*}[t]
	\centering
	\scriptsize
	
	\subfloat[ORCA]
	{
		\includegraphics[width=0.23\textwidth]{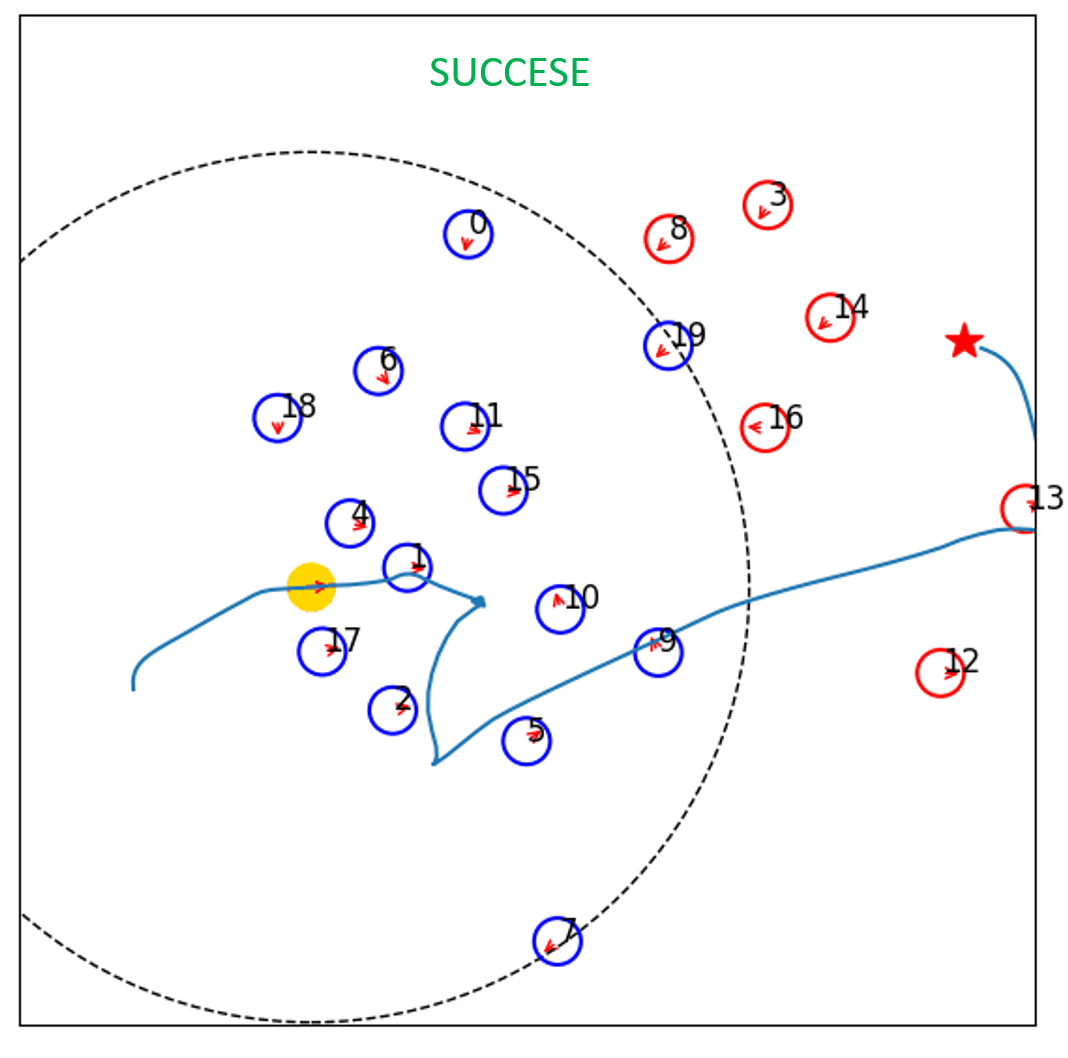}
	}
	\subfloat[SFM]
	{
		\includegraphics[width=0.23\textwidth]{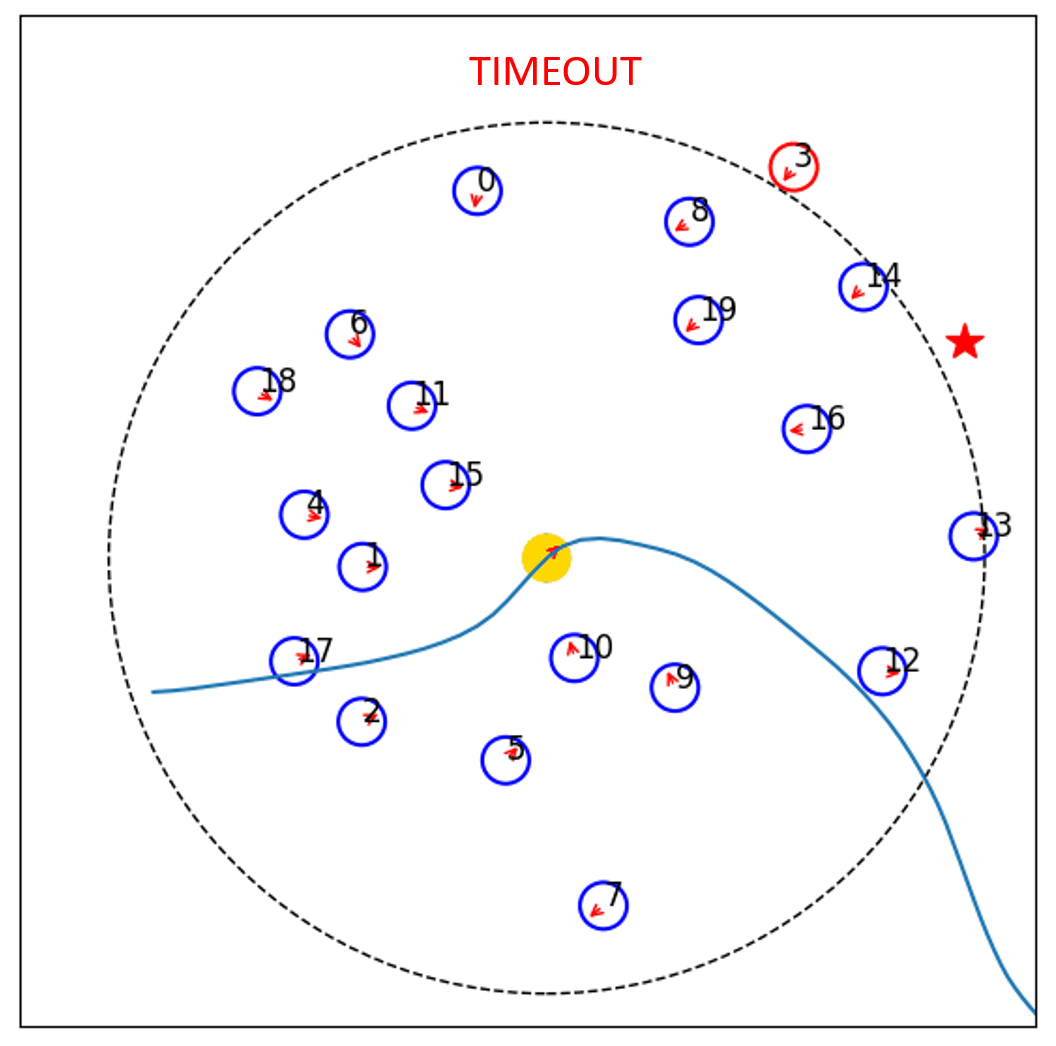}
	}
	\subfloat[IntentionGRU]
	{
		\includegraphics[width=0.23\textwidth]{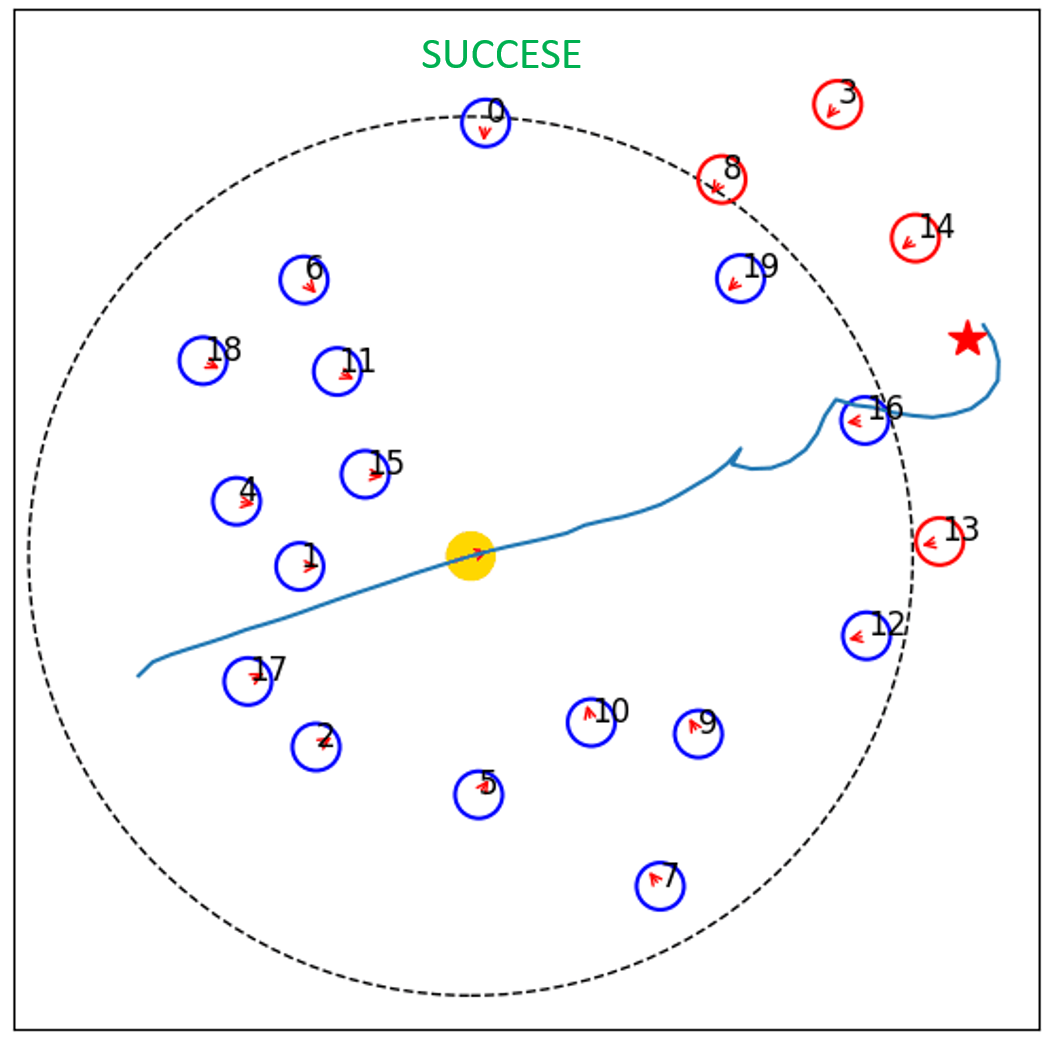}
	}
	\subfloat[IntentionGRU\_Traj(OURS)]
	{
		\includegraphics[width=0.23\textwidth]{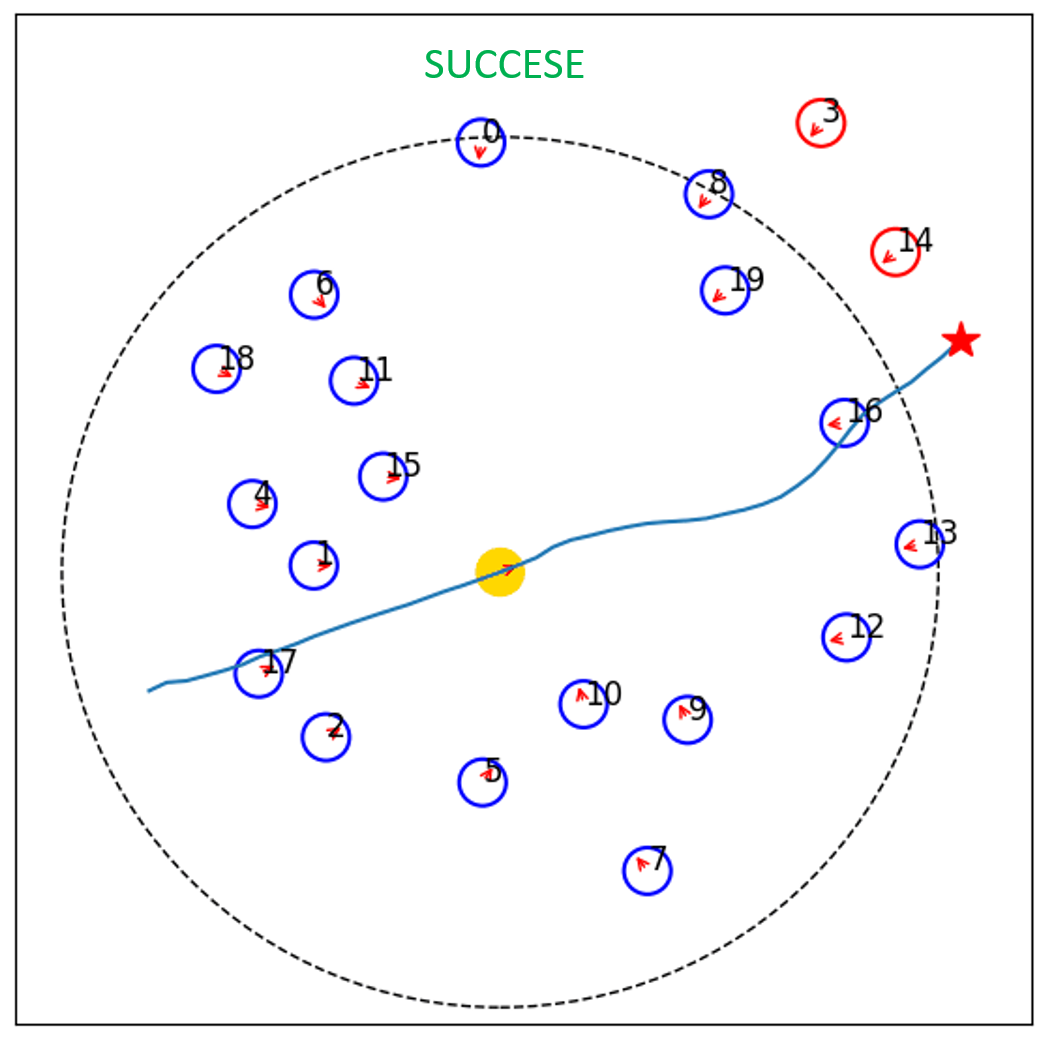}
	}
	\captionsetup{font=small}
	\caption{The comparison of navigation trajectories in high-density circle crossing.}
	\label{fig:traj_comp}
\end{figure*}

A qualitative analysis was also conducted for ORCA, SFM, IntentionGRU, and IntentionGRU\_Traj using the same high-density configuration, with identical start and goal locations and identical pedestrian trajectories. The resulting path comparisons are presented in Fig.~\ref{fig:traj_comp}. As shown, SFM fails due to a timeout, while ORCA produces a relatively inefficient and winding path. IntentionGRU yields a generally smooth trajectory but exhibits abrupt directional changes near the end as a result of emergency obstacle avoidance. In contrast, IntentionGRU\_Traj produces a noticeably smoother and more efficient path, clearly demonstrating its superior overall performance relative to the other methods.

\subsection{Verification in the 3D Scenario}
\label{sec:5.2.5}

This 3D experiment is conducted in the DRL-VO simulation environment, where AMCL provides localization and the SFM model simulates pedestrian motion. The 2D strategy trained in this study is used as a local planner for obstacle avoidance. Overall, the results demonstrate the preliminary feasibility of deploying the proposed method. For further details, please refer to the link \url{https://www.youtube.com/watch?v=R9jizMgak1E}.

\section{CONCLUSION, DISCUSSION AND Limitations}
\label{sec:5}
This work presents a trajectory-oriented framework for enhancing crowd navigation. A priority-based multi-objective evaluation system, a curvature-based $C^2$ continuity metric, and a density-aware reward strategy are introduced to improve trajectory smoothness while maintaining safety and efficiency. Experiments across different densities show consistent performance gains, and 3D tests confirm preliminary deployability. Nonetheless, transferring 2D-trained policies to 3D settings may introduce observation bias and widen the sim-to-real gap. Future work will explore training in realistic 3D simulators and employing end-to-end models to mitigate cumulative errors.

\bibliographystyle{IEEEtran}
\bibliography{IEEEabrv,bibfile}
\end{document}